\documentclass[runningheads]{llncs}
\usepackage{graphicx}
\usepackage{tabularx,ragged2e}
\usepackage{booktabs}
\usepackage{amsmath,amssymb,amsfonts} %
\usepackage{calligra}
\usepackage{bbm}
\usepackage{color}
\usepackage[width=122mm,left=12mm,paperwidth=146mm,height=193mm,top=12mm,paperheight=217mm]{geometry}
\usepackage{appendix}

\usepackage{epstopdf}

\usepackage{times}
\usepackage{epsfig}
\usepackage{dsfont}
\usepackage{siunitx}
\usepackage{subfigure}
\usepackage{multirow}
\usepackage{caption}
\usepackage{booktabs}       %
\usepackage{standalone}
\usepackage{xspace}
\usepackage{comment}
\usepackage{hyperref}
\usepackage{harpoon}
\usepackage{lipsum}

\newcommand{\myparagraph}[1]{\noindent{\bf #1}}

\newcommand{\anja}[1]{\textcolor{green}{Anja: #1}}
\newcommand\numberthis{\addtocounter{equation}{1}\tag{\theequation}}

\newcommand{\modelname}{Equalizer}
\newcommand{\lossnameapp}{Appearance Confusion Loss}
\newcommand{\lossnamecon}{Confident Loss}

\newcommand\blfootnote[1]{%
  \begingroup
  \renewcommand\thefootnote{}\footnote{#1}%
  \addtocounter{footnote}{-1}%
  \endgroup
}

\begin{document}
\pagestyle{headings}
\mainmatter

\title{Women also Snowboard: \\
Overcoming Bias in Captioning Models
} %

\author{Kaylee Burns*$^1$, Lisa Anne Hendricks*$^1$, Kate Saenko$^2$,\\ Trevor Darrell$^1$, Anna Rohrbach$^1$}
\institute{$^1$ UC Berkeley $^2$ Boston University}

\maketitle

\begin{abstract}
Most machine learning methods are known to capture and exploit biases of the training data. 
While some biases are beneficial for learning, others are harmful. Specifically, image captioning models tend to exaggerate biases present in training data (e.g., if a word is present in 60\% of training sentences, it might be predicted in 70\% of sentences at test time). This can lead to incorrect captions in domains where unbiased captions are desired, or required, due to over-reliance on the learned prior and image context.  In this work we investigate generation of gender-specific caption words (e.g. man, woman) based on the person's appearance or the image context. We introduce a new \emph{\modelname{}} model that encourages equal gender probability when gender evidence is occluded in a scene and confident predictions when gender evidence is present. The resulting model is forced to look at a person rather than use contextual cues to make a gender-specific prediction. The losses that comprise our model, the \emph{\lossnameapp{}} and the \emph{\lossnamecon{}},  are general, and can be added to any description model in order to mitigate impacts of unwanted bias in a description dataset. Our proposed model has lower error than prior work when describing images with people and mentioning their gender and more closely matches the ground truth ratio of sentences including women to sentences including men. Finally, we show that 
our model more often looks at people when predicting their gender. \footnote{\url{https://people.eecs.berkeley.edu/~lisa\_anne/snowboard.html}}

\blfootnote{* Authors contributed equally, listed alphabetically.} 
\end{abstract}

\section{Introduction}

\begin{figure}[t]
\centering
\includegraphics[width=\linewidth]{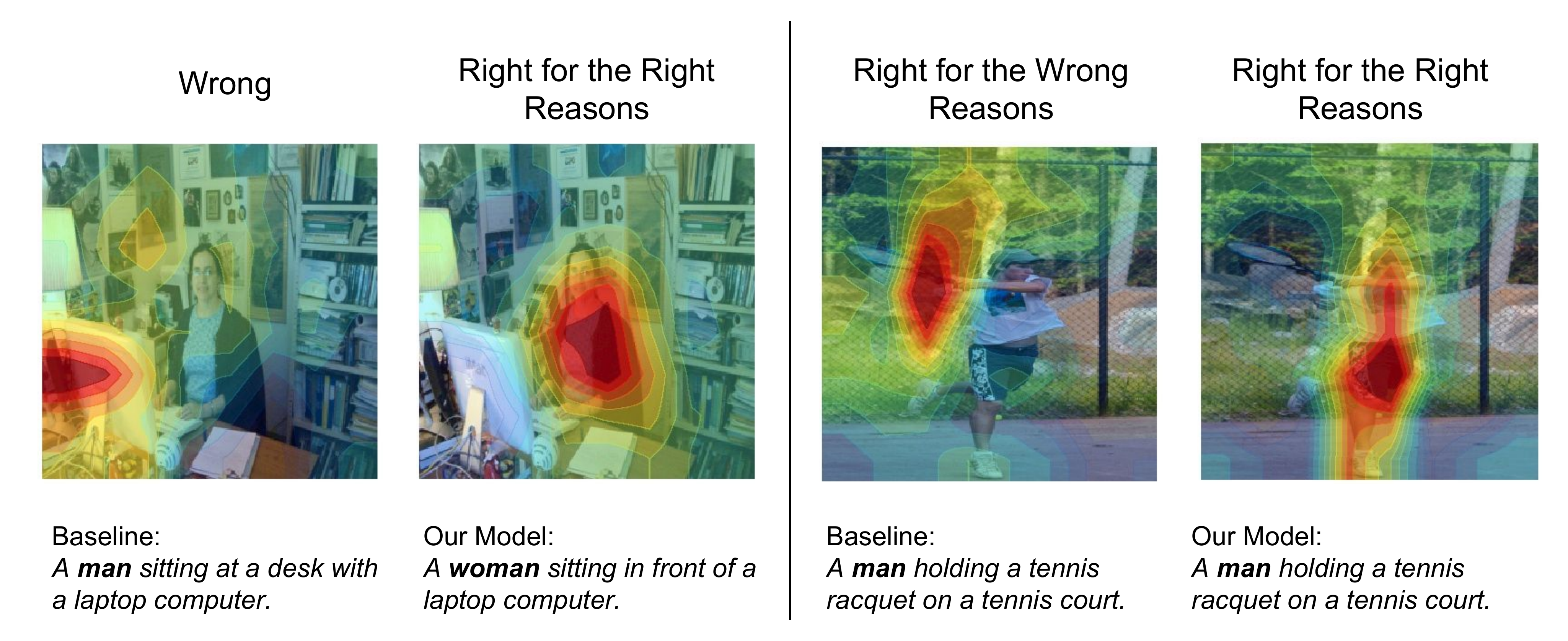}
\caption{\label{fig:teaser}  
Examples where our proposed model (\modelname{}) corrects bias in image captions. The overlaid heatmap indicates which image regions are most important for predicting the gender word.
On the left, the baseline predicts gender incorrectly, presumably because it looks at the laptop (not the person).  %
On the right, the baseline predicts the gender correctly but it 
does not look at the person when predicting gender and is thus not acceptable.  In contrast, our model predicts the correct gender word and correctly considers the person when predicting gender.}
\end{figure}

Exploiting contextual cues can frequently lead to better performance on computer vision tasks~\cite{torralba2001statistical,torralba2002contextual,gkioxari2015contextual}.
For example, in the visual description task, predicting a ``mouse'' might be easier given that a computer is also in the image. 
However, in some cases making decisions based on context can lead to incorrect, and perhaps even offensive, predictions.
In this work, we consider one such scenario: generating captions about men and women.
We posit that when description models predict gendered words such as ``man'' or ``woman'', they should consider visual evidence associated with the described person, and not contextual cues like location (e.g., ``kitchen'') or other objects in a scene (e.g., ``snowboard'').
Not only is it important for description systems to avoid egregious errors (e.g., always predicting the word ``man'' in snowboarding scenes), but it is also important for predictions to be right for the right reason.  For example, Figure~\ref{fig:teaser} (left) shows a case where prior work predicts the incorrect gender, while our model accurately predicts the gender by considering the correct gender evidence.
Figure~\ref{fig:teaser} (right) shows an example where both models predict the correct gender, but prior work does not look at the person when describing the image (it is right for the wrong reasons).

Bias in image captioning is particularly challenging to overcome because of the multimodal nature of the task; predicted words are not only influenced by an image, but also biased by the learned language model. 
Though \cite{zhao2017men} studied bias for structured prediction tasks (e.g., semantic role labeling), they did not consider the task of image captioning.
Furthermore, the solution proposed in \cite{zhao2017men} requires access to the entire test set in order to rebalance gender predictions to reflect the distribution in the training set.
Consequently, \cite{zhao2017men} relies on the assumption that the distribution of genders is the same at training and test time.
We make no such assumptions; we consider a more realistic scenario in which captions are generated for images independent of other test images.

In order to encourage description models to generate less biased captions, we introduce the \textit{\modelname{}} Model.   
Our model includes two complementary loss terms: the \textit{\lossnameapp{} (ACL)} and the \textit{\lossnamecon{} (Conf)}.
The \lossnameapp{} is based on the intuition that, given an image in which evidence of gender is absent, description models should be unable to accurately predict a gendered word.
However, it is not enough to confuse the model when gender evidence is absent; we must also encourage the model to consider gender evidence when it is present.
Our \lossnamecon{} helps to increase the model's confidence when gender is in the image.
These complementary losses allow the \modelname{} model to be cautious in the absence of gender information and discriminative in its presence. %

Our proposed \modelname{} model leads to less biased captions: not only does it lead to lower error when predicting gendered words, but it also performs well when the distribution of genders in the test set is not aligned with the training set.
Additionally, we observe that \modelname{} generates gender neutral words (like ``person'') when it is not confident of the gender.
Furthermore, we demonstrate that \modelname{} focuses on humans when predicting gender words, as opposed to focusing on other image context.

\section{Related Work}

\myparagraph{Unwanted Dataset Bias.} %
Unwanted dataset biases (e.g., gender, ethnic biases) have been studied across a wide variety of AI domains~\cite{ryu2017improving,stock2017convnets,bolukbasi2016man,buolamwini2017gender,barocas2016big,united2014big}.
One common theme is the notion of \textit{bias amplification}, in which bias is not only learned, but amplified~\cite{zhao2017men,bolukbasi2016man,stock2017convnets}.
For example, in the image captioning scenario, if 70\% of images with umbrellas include a woman and 30\% include a man, at test time the model might amplify this bias to 85\% and 15\%. 
Eliminating bias amplification is not as simple as balancing across attributes for a specific category.
~\cite{stock2017convnets} study bias in classification and find that even though white and black people appear in ``basketball'' images with similar frequency, models learn to classify images as ``basketball'' based on the presence of a black person.
One explanation is that though the data is balanced in regard to the class ``basketball'', there are many more white people in the dataset.
Consequently, to perfectly balance a dataset, one would have to balance across all possible co-occurrences which is infeasible.

Natural language data is subject to \textit{reporting bias} \cite{bolukbasi2016man,gordon13acmw,misra2016seeing,van2016stereotyping} in which people over-report less common co-occurrences, such as ``male nurse''~\cite{bolukbasi2016man} or ``green banana''~\cite{misra2016seeing}.
\cite{van2016stereotyping} also discuss how visual descriptions reflect cultural biases (e.g., assuming a woman with a child is a mother, even though this cannot be confirmed in an image).
We observe that annotators specify gender even when gender cannot be confirmed in an image (e.g., a snowboarder might be labeled as ``man'' even if gender evidence is occluded).%

Our work is most similar to \cite{zhao2017men} who consider bias in semantic role labeling and multilabel classification (as opposed to image captioning).
To avoid bias amplification, \cite{zhao2017men} rebalance the test time predictions to more accurately reflect the training time word ratios.
This solution is unsatisfactory because (i) it requires access to the entire test set and (ii) it assumes that the distribution of objects at test time is the same as at training time.
We consider a more realistic scenario in our experiments, and show that the ratio of woman to man in our predicted sentences closely resembles the ratio in ground truth sentences, even when the test distribution is different from the training distribution.

\myparagraph{Fairness.} Building AI systems which treat \textit{protected attributes} (e.g., age, gender, sexual orientation) in a fair manner is increasingly important%
~\cite{hardt2016equality,dwork2012fairness,zhang2018mitigating,quadrianto2017recycling}.
In the machine learning literature, ``fairness'' generally requires that systems do not use information such as gender or age in a way that disadvantages one group over another.
We consider is different scenario %
as we are trying to \textit{predict} protected attributes.

\textit{Distribution matching} has been used to build fair systems~\cite{quadrianto2017recycling} by encouraging the distribution of decisions to be similar across different protected classes, as well as for other applications such as domain adaption~\cite{tzeng2015simultaneous,zhang2015deep} and transduction learning~\cite{quadrianto2009distribution}.
Our \lossnameapp{} is similar as it encourages the distribution of predictions to be similar for man and woman classes when gender information is not available.

\myparagraph{Right for the Right Reasons.}  %
Assuring models are ``right for the right reasons,'' or consider similar evidence as humans when making decisions, helps researchers understand how models will perform in real world applications (e.g., when predicting outcomes for pneumonia patients in~\cite{caruana2015intelligible}) or discover underlying dataset bias~\cite{tan2017detecting}.
We hypothesize that models which look at appropriate gender evidence will perform better in new scenarios, specifically when the gender distribution at test and training time are different. %

Recently,~\cite{ross2017right} develop a loss function which compares explanations for a decision to ground truth explanations.
However, ~\cite{ross2017right} generating explanations for visual decisions is a difficult and active area of research
~\cite{ramanishka2017top,selvaraju2016grad,fong2017interpretable,ribeiro2016should,zintgraf2017visualizing,zeiler2014visualizing}.
Instead of relying on our model to accurately explain itself during training, we verify that our formulation encourages models to be right for the right reason at test time.

\myparagraph{Visual Description.}  Most visual description work (e.g., \cite{vinyals2015show,donahue2015long,karpathy2015deep,xu2015show,anderson2017bottom}) focuses on improving overall sentence quality, without regard to captured biases.
Though we pay special attention to gender in this work, all captioning models trained on visual description data (MSCOCO~\cite{lin2014microsoft}, Flickr30k~\cite{young2014image}, MSR-VTT~\cite{xu2016msr} to name a few) implicitly learn to classify gender.
However current captioning models do not discuss gender the way humans do, but \textit{amplify} gender bias; our intent is to generate descriptions which more accurately reflect human descriptions when discussing this important category.

\myparagraph{Gender Classification.}  
Gender classification models frequently focus on facial features~\cite{levi2015age,zhang2016gender,eidinger2014age}.
In contrast, we are mainly concerned about whether contextual clues in complex scenes bias the production of gendered words during sentence generation.
Gender classification has also been studied in natural language processing (\cite{argamon2007mining,yan2006gender}, \cite{burger2011discriminating}).

\myparagraph{Ethical Considerations.}  
Frequently, gender classification is seen as a binary task: data points are labeled as either ``man'' or ``woman''.
However, AI practitioners, both in industrial\footnote{\scriptsize{{https://clarifai.com/blog/socially-responsible-pixels-a-look-inside-clarifais-new-demographics-recognition-model}}} and academic\footnote{\scriptsize{{https://www.media.mit.edu/projects/gender-shades/faq}}} settings, are increasingly concerned that gender classification systems should be inclusive. 
Our captioning model predicts three gender categories: male, female, and gender neutral (e.g., person) based on visual appearance. 
When designing gender classification systems, it is important to understand where labels are sourced from~\cite{larson2017gender}.
We determine gender labels using a previously collected publicly released dataset in which annotators describe images \cite{lin2014microsoft}.
Importantly, people in the images are not asked to identify their gender.  
Thus, we emphasize that we are not classifying biological sex or gender identity, but rather outward gender appearance.%

\section{\modelname{}: Overcoming Bias in Description Models}

\begin{figure}[t]
\centering
\includegraphics[width=0.9\linewidth]{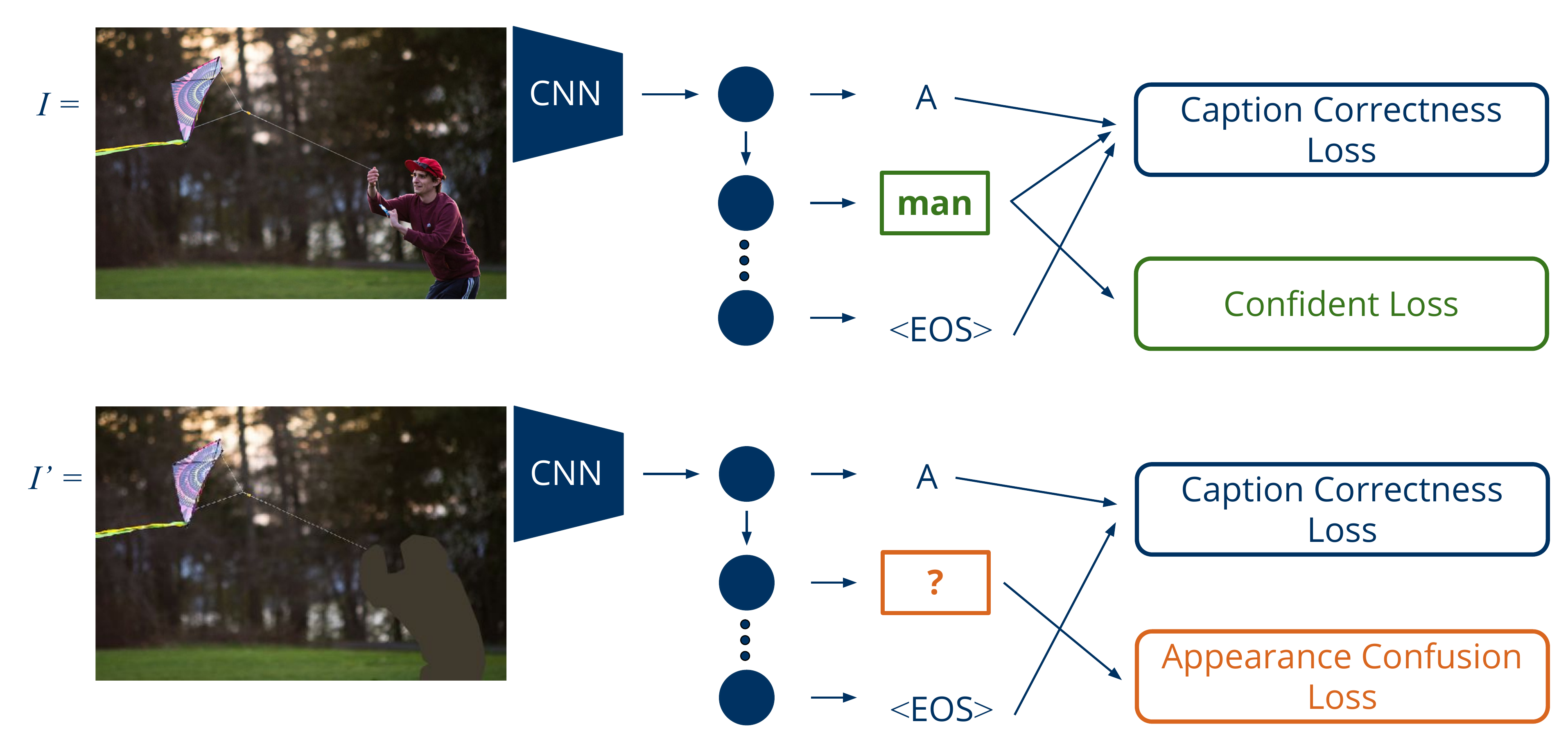}
\caption{\modelname{} includes two novel loss terms: the \lossnamecon{} on images with men or women (top) and the \lossnameapp{} on images where men and women are occluded (bottom). Together these losses encourage our model to make correct predictions when evidence of gender is present, and be cautious in its absence. We also include the Caption Correctness Loss (cross entropy loss) for both image types.}
\label{fig:model}
\end{figure}

\modelname{} is based on the following intuitions: if evidence to support a specific gender decision is not present in an image, the model should be \textit{confused} about which gender to predict (enforced by an \lossnameapp{} term), and if evidence to support a gender decision is in an image, the model should be \textit{confident} in its prediction (enforced by a \lossnamecon{} term). 
To train our model we require not only pairs of images, $I$, and sentences, $S$, but also annotation masks $M$ which indicate which evidence in an image is appropriate for determining gender.
Though we use~\cite{vinyals2015show} as our base network, \modelname{} is general and can be integrated into any deep description frameworks.

\subsection{Background: Description Framework}

To generate a description, high level image features are first extracted from the InceptionV3 \cite{szegedy2016rethinking} model.
The image features are then used to initialize an LSTM hidden state.
To begin sentence generation, a start of sentence token is input into the LSTM.
For each subsequent time step during training, the ground truth word $w_t$ is input into the LSTM.  
At test time, the previously predicted word $w_{t-1}$ is input into the LSTM at each time step.
Generation concludes when an end of sequence token is generated.
Like~\cite{vinyals2015show}, we include the standard cross entropy loss ($\mathcal{L}^{CE}$) during training:
\begin{equation}
\mathcal{L}^{CE} = - \frac{1}{N} \sum_{n=0}^N \sum_{t=0}^T \text{log} (p(w_t|w_{0:t-1}, I)),
\end{equation}
where $N$ is the batch size, $T$ is the number of words in the sentence, $w_t$ is a ground truth word at time $t$, and $I$ is an image.

\subsection{\lossnameapp{}}

Our \lossnameapp{} encourages the underlying description model to be \textit{confused} when making gender decisions if the input image does not contain appropriate evidence for the decision. To optimize the \lossnameapp{}, we require ground truth rationales indicating which evidence is appropriate for a particular gender decision.
We expect the resulting rationales to be masks, $M$, which are $1$ for pixels which should not contribute to a gender decision and $0$ for pixels which are appropriate to consider when determining gender.
The Hadamard product of the mask and the original image, $I \odot M$, yields a new image, $I'$, with gender information that the implementer deems appropriate for classification removed. %
Intuitively, for an image devoid of gender information, %
the probability of predicting man or woman should be equal. 
The \lossnameapp{} enforces a fair prior by asserting that this is the case.%

To define our \lossnameapp{}, we first define a \textit{confusion} function ($\mathcal{C}$) which operates over the predicted distribution of words $p(\tilde{w}_t)$, a set of woman gender words ($\mathcal{G}_w$), and a set of man gender words ($\mathcal{G}_m$):
\begin{align*}
\mathcal{C}(\tilde{w}_t, I') = |\sum_{g_w \in \mathcal{G}_w} p(\tilde{w}_t=g_w|w_{0:t-1}, I')  - \sum_{g_m \in \mathcal{G}_m} p(\tilde{w}_t = g_m|w_{0:t-1}, I')| \numberthis \label{eqn}.
\end{align*}

In practice, the $\mathcal{G}_w$ consists only of the word ``woman'' and, likewise, the $\mathcal{G}_m$ consists only of the word ``man''.
These are by far the most commonly used gender words in the datasets we consider and we find that using these ``sets'' results in similar performance as using more complete sets. %

We can now define our \lossnameapp{} ($\mathcal{L}^{AC}$) as:
\begin{equation}
\mathcal{L}^{AC} = \frac{1}{N} \sum_{n=0}^N \sum_{t=0}^T \mathbbm{1}(w_t \in \mathcal{G}_w \cup \mathcal{G}_m) \mathcal{C}(\tilde{w}_t, I'),
\end{equation}
where $\mathbbm{1}$ is an indicator variable that denotes whether or not $w_t$ is a gendered word.

For the remaining non-gendered words that correspond to images $I'$, we apply the standard cross entropy loss   %
to encourage the model to discuss objects which are still visible in $I'$. In addition to encouraging sentences to be image relevant even when the gender information has been removed, this also encourages the model to learn representations of words like ``dog'' and ``frisbee'' that are not reliant on gender information.

\subsection{\lossnamecon{}}

In addition to being unsure when gender evidence is occluded, we also encourage our model to be confident when gender evidence is present.
Thus, we introduce the \lossnamecon{} term, which encourages the model to predict gender words correctly.  
Our \lossnamecon{} encourages the probabilities for predicted gender words to be high on images $I$ in which gender information is present. 
Given functions $\mathcal{F}^W$ and $\mathcal{F}^M$ which measure how confidently the model predicts woman and man words respectively,
we can write the \lossnamecon{} as:

\begin{equation}
\mathcal{L}^{Con} = \frac{1}{N} \sum_{n=0}^N \sum_{t=0}^T (\mathbbm{1}(w_t \in \mathcal{G}_w) \mathcal{F}^W (\tilde{w_t}, I) + \mathbbm{1}(w_t \in \mathcal{G}_m) \mathcal{F}^M (\tilde{w_t}, I)).
\end{equation}

To measure the confidence of predicted gender words, we consider the quotient between predicted probabilities for man and gender words ($\mathcal{F}^M$ is of the same form):
\begin{equation}
\mathcal{F}^W (\tilde{w_t}, I)= \frac{\sum_{g_m \in \mathcal{G}_m} p(\tilde{w_t}= g_m|w_{0:t-1}, I)}{(\sum_{g_w \in \mathcal{G}_w} p(\tilde{w_t} = g_w|w_{0:t-1}, I)) + \epsilon}
\end{equation}
where $\epsilon$ is a small epsilon value added for numerical stability. 

When the model is confident of a gender prediction (e.g., for the word ``woman''), the probability of the word ``woman'' should be considerably higher than the probability of the word ``man'', which will result in a small value for $\mathcal{F}^W$ and thus a small loss.
One nice property of considering the quotient between predicted probabilities is that we encourage the model to distinguish between gendered words without forcing the model to predict a gendered word.
For example, if the model predicts a probability of $0.2$ for ``man'', $0.5$ for ``woman'', and $0.3$ for ``person'' on a ``woman'' image, our confidence loss will be low.
However, the model is still able to predict gender neutral words, like ``person'' with relatively high probability.
This is distinct from other possible losses, like placing a larger weight on gender words in the cross entropy loss, which forces the model to predict ``man''/``woman'' words and penalizes the gender neutral words.

\subsection{The \modelname{} Model}

Our final model is a linear combination of all aforementioned losses:
\begin{equation}
\mathcal{L} = \alpha\mathcal{L}^{CE} + \beta\mathcal{L}^{AC} + \mu\mathcal{L}^{Con},
\end{equation}
where $\alpha$, $\beta$, and $\mu$ are hyperparameters chosen on a validation set ($\alpha, \mu = 1$, $\beta = 10$ in our experiments). %

Our \modelname{} method is general and our base captioning framework can be substituted with any other deep captioning framework.
By combining all of these terms, the \modelname{} model can not only generate image relevant sentences, but also make confident gender predictions under sufficient evidence. We find that both the \lossnameapp{} and the \lossnamecon{} are important in creating a confident yet cautious model. Interestingly, the \modelname{} model achieves the lowest misclassification rate only when these two losses are combined, highlighting the complementary nature of these two loss terms.

\section{Experiments}

\subsection{Datasets}

\myparagraph{MSCOCO-Bias.} To evaluate our method, we consider the dataset used by \cite{zhao2017men} for evaluating bias amplification in structured prediction problems.
This dataset consists of images from MSCOCO~\cite{lin2014microsoft} which are labeled as ``man'' or ``woman''.  
Though ``person'' is an MSCOCO class, ``man'' and ``woman'' are not, so \cite{zhao2017men} employ ground truth captions to determine if images contain a man or a woman. 
Images are labeled as ``man'' if at least one description includes the word ``man'' and no descriptions include the word ``woman''.
Likewise, images are labeled as ``woman'' if at least one description includes the word ``woman'' and no descriptions include the word ``man''.
Images are discarded if both ``man'' and ``woman'' are mentioned.
We refer to this dataset as MSCOCO-Bias.

\myparagraph{MSCOCO-Balanced.}  We also evaluate on a set where we purposely change the gender ratio.
We believe this is representative of real world scenarios in which different distributions of men and women might be present at test time.
The MSCOCO-Bias set has a roughly 1:3 woman to man ratio where as this set, called MSCOCO-Balanced, has a 1:1 woman to man ratio. 
We randomly select 500 images from MSCOCO-Bias set which include the word ``woman'' and 500 which include ``man''.

\myparagraph{Person Masks.}  To train \modelname{}, we need ground truth human rationales for why a person should be predicted as a man or a woman.
We use the person segmentation masks from the MSCOCO dataset. Once the masked image is created, we fill the segmentation mask with the average pixel value in the image. 
We use the masks both at training time to compute \lossnameapp{} and during evaluation to ensure that models are predicting gender words by looking at the person. While for MSCOCO the person annotations are readily available, for other datasets e.g. a person detector could be used.

\subsection{Metrics}

To evaluate our methods, we rely on the following metrics. 

\textbf{Error.} %
Due to the sensitive nature of prediction for protected classes (gender words in our scenario), we emphasize the importance of a low error. The error rate is the number of man/woman misclassifications, while gender neutral terms are not considered errors. We expect that the best model would rather predict gender neutral words in cases where gender is not obvious.

\textbf{Gender Ratio.} Second, we consider the ratio of sentences which belong to a ``woman'' set to sentences which belong to a ``man'' set. We consider a sentence to fall in a ``woman'' set if it predicts any word from a precompiled list of female gendered words, and respectively fall in a ``man'' set if it predicts any word from a precompiled list of male gendered words. %

\textbf{Right for the Right Reasons.} Finally, to measure if a model is ``right for the right reasons'' we consider the pointing game \cite{zhang2016top} evaluation. We first create visual explanations for ``woman''/``man'' using the Grad-CAM approach~\cite{selvaraju2016grad} as well as saliency maps created by occluding image regions in a sliding window fashion. To measure if our models are right for the right reason, we verify whether the point with the highest activation in the explanation heat map falls in the person segmentation mask.

\subsection{Training Details} All models are initialized from the Show and Tell model \cite{vinyals2015show} pre-trained on all of MSCOCO for 1 million iterations (without fine-tuning through the visual representation). 
Models are trained for additional 500,000 iterations on the MSCOCO-Bias set,   %
fine-tuning through the visual representation (Inception v3 \cite{szegedy2016rethinking}) for 500,000 iterations.

\subsection{Baselines and Ablations}

\myparagraph{Baseline-FT.} The simplest baseline is fine-tuning the Show and Tell model through the LSTM and convolutional networks using the standard cross-entropy loss on our target dataset, the MSCOCO-Bias dataset.  

\myparagraph{Balanced.} %
We train a Balanced baseline in which we re-balance the data distribution at training time to account for the larger number of men instances in the training data. Even though we cannot know the correct distribution of our data at test time, we can enforce our belief that predicting a woman or man should be equally likely.
At training time, we re-sample the images of women so that the number of training examples of women is the same as the number of training examples of men. %

\myparagraph{UpWeight.} 
We also experiment with upweighting the loss value for gender words in the standard cross entropy loss to increase the penalty for a misclassification.  %
For each time step where the ground truth caption says the word  ``man'' or ``woman'', we multiply that term in the loss by a constant value (10 in reported experiments).
Intuitively, upweighting should encourage the models to accurately predict gender words. 
However, unlike our \lossnamecon{}, upweighting drives the model to make either ``man'' or ``woman'' predictions without the opportunity to place a high probability on 
gender neutral words.

\myparagraph{Ablations.} To isolate the impact of the two loss terms in \modelname{}, we report results with only the Appearance Confusion Loss (\modelname{} w/o Conf) and only the Confidence Loss (\modelname{} w/o ACL).
We then report results of our full \modelname{} model.

\subsection{Results}

\begin{table*}[t]
\centering
\begin{tabular}{ l@{\ \ }| r r | r r }
& \multicolumn{2} {c@{\ \ }@{\ \ }|} {MSCOCO-Bias}  & \multicolumn{2} {c} {MSCOCO-Balanced} \\
Model & Error & Ratio $\Delta$  & Error &  Ratio $\Delta$  \\
\midrule
  \midrule
Baseline-FT  & 12.83   & 0.15  & 19.30 &  0.51  \\
Balanced    & 12.85   &  0.14  & 18.30 &  0.47  \\
UpWeight    & 13.56  & 0.08  & 16.30 &  0.35  \\
\midrule
\modelname{} w/o ACL  & 7.57 & 0.04 &    10.10 & 0.26    \\
\modelname{} w/o Conf  & 9.62 & 0.09 & 13.90 & 0.40    \\
\modelname{} &  \textbf{7.02} & \textbf{-0.03}   & \textbf{8.10} & \textbf{0.13}   \\
\end{tabular}
\caption{\label{tab:accuracy-ratio} Evaluation of predicted gender words based on error rate and ratio of generated sentences which include the ``woman'' words to sentences which include the ``man'' words. \modelname{} achieves the lowest error rate and predicts sentences with a gender ratio most similar to the corresponding ground truth captions (Ratio $\Delta$), even when the test set has a different distribution of gender words than the training set, as is the case for the MSCOCO-Balanced dataset.}%
\end{table*}

\myparagraph{Error.} Table~\ref{tab:accuracy-ratio} reports the error rates when describing men and women %
on the MSCOCO-Bias and MSCOCO-Balanced test sets.
Comparing to baselines, \modelname{} shows consistent improvements.
Importantly, our full model consistently improves upon \modelname{} w/o ACL and \modelname{} w/o Conf.
When comparing \modelname{} to baselines, we see a larger performance gain on the MSCOCO-Balanced dataset.
As discussed later, this is in part because our model does a particularly good job of decreasing error on the minority class (woman).
Unlike baseline models, our model has a similar error rate on each set.
This indicates that the error rate of our model is not as sensitive to shifts in the gender distribution at test time.

Interestingly, the results of the Baseline-FT model and Balanced model are not substantially different.
One possibility is that the co-occurrences across words are not balanced (e.g., if there is gender imbalance specifically for images with ``umbrella'' just balancing the dataset based on gender word counts is not sufficient to balance the dataset).  We emphasize that balancing across all co-occurring words is difficult in large-scale settings with large vocabularies.

\myparagraph{Gender Ratio}  We also consider the ratio of captions which include only female words to captions which include only male words. 
In Table~\ref{tab:accuracy-ratio} we report the \textit{difference} between the ground truth ratio and the ratio produced by each captioning model.
Impressively, \modelname{} achieves the closest ratio to ground truth on both datasets. %
Again, the ACL and Confident losses are complementary and \modelname{} has the best overall performance.

\begin{table*}[t]
\centering
\begin{tabular}{l@{\ \ }|@{\ \ }ccc@{\ \ }|@{\ \ }ccc|c}
&& Women &&& Men & & Outcome Divergence\\
Model & Correct & Incorrect & Other & Correct & Incorrect & Other & between Genders\\ \midrule
Baseline-FT  &   46.28 &	34.11 &	19.61 &	75.05 &	4.23	& 20.72  & 0.121 \\
Balanced     &   47.67 &	33.80 &	18.54 &	75.89 &	4.38 &	19.72  &  0.116 \\
UpWeight     &    \textbf{60.59} &	29.82 &	9.58 &	\textbf{87.84} &	6.98 &	5.17 & 0.078 \\
\midrule
\modelname{} w/o ACL  & 56.18 &	16.02 &	27.81 &	67.58 &	\textbf{4.15} &	28.26  & 0.031 \\
\modelname{} w/o Conf &   46.03 &	24.84  & 29.13  & 61.11 &  3.47  &  35.42  & 0.075 \\
\modelname{} (Ours) &   57.38 &	\textbf{12.99} &	29.63 &	59.02 &	4.61 &	36.37  & \textbf{0.018} \\
\end{tabular}
\caption{\label{tab:results_ind_gender}Accuracy per class for MSCOCO-Bias dataset. Though UpWeight achieves the highest recall for both men and women images, it also has a high error, especially for women. One criterion of a ``fair'' system is that it has similar outcomes across classes. We measure outcome similarity by computing the Jensen-Shannon divergence between Correct/Incorrect/Other sentences for men and women images (lower is better) and observe that Equalizer performs best on this metric.   %
}
\end{table*}

%
%

%

%
%

%
%
%
%
%
%
%
%
%
%
%
%
%
%
%
%
%
%

%
%
%
%
%

\myparagraph{Performance for Each Gender.} 
Images with females comprise a much smaller portion of MSCOCO than images with males. Therefore the overall performance across  classes (i.e. man, woman) can be misleading because it downplays the errors in the minority class.
Additionally, unlike~\cite{zhao2017men} who consider a classification scenario in which the model is forced to predict a gender, our description models can also discuss gender neutral terms such as ``person'' or ``player''. 
In Table~\ref{tab:results_ind_gender} for each gender, we report the percentage of sentences in which gender is predicted correctly or incorrectly and when no gender specific word is generated on the MSCOCO-Bias set.

Across all models, the error for Men is quite low. 
However, our model significantly improves the error for the minority class, Women. 
Interestingly, we observe that Equalizer has a similar recall (Correct), error (Incorrect), and Other rate across both genders.
A caption model could be considered more ``fair'' if, for each gender, the possible outcomes (correct gender mentioned, incorrect gender mentioned, gender neutral) are similar.
This resembles the notion of equalized odds in fairness literature~\cite{hardt2016equality}, which requires a system to have similar false positive and false negative rates across groups.
To formalize this notion of fairness in our captioning systems, we report the outcome type divergence between genders by measuring the Jensen-Shannon~\cite{lin1991divergence} divergence between Correct/Incorrect/Other outcomes for Men and Women. 
Lower divergence indicates that Women and Men classes result in a similar distribution of outcomes, and thus the model can be considered more ``fair''.
Equalizer has the lowest divergence ($0.018$).

\begin{figure}[t]
\centering
\includegraphics[width=0.8\linewidth]{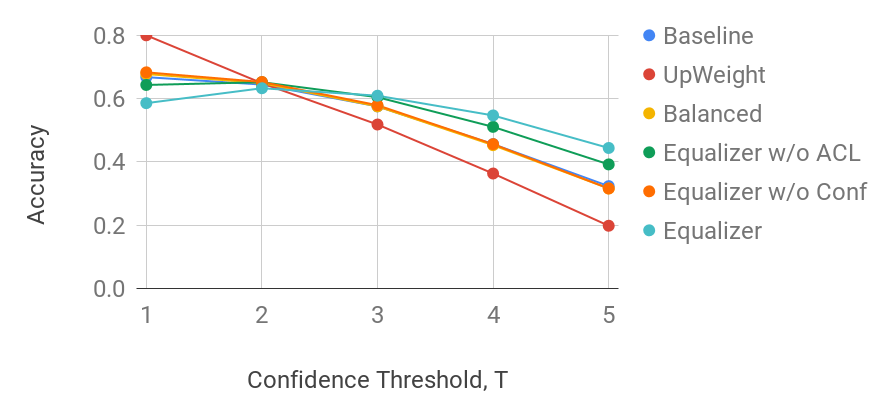}
\caption{\label{fig:confident_accuracy} Accuracy across man, woman, and gender neutral terms for different models as a function of annotator confidence.  When only one annotator describes an image with a gendered word, Equalizer has a low accuracy as it more likely predicts gender neutral words but when more annotations mention gendered words, Equalizer has higher accuracy than other models.}
\end{figure}

\myparagraph{Annotator Confidence.} %
As described above, gender labels are mined from captions provided in the MSCOCO dataset.
Each image corresponds to five captions, but not all captions for a single image include a gendered word. 
Counting the number of sentences which include a gendered word provides a rough estimate of how apparent gender is in an image and how important it is to mention when describing the scene.

To understand how well our model captures the way annotators describe people,   
instead of labeling images as either ``man'' or ``woman'', we label images as ``man'', ``woman'', or ``gender neutral'' based on how many annotators mentioned gender in their description. %
For a specific threshold value $T$, we consider an image to belong to the ``man'' or ``woman'' class if $T$ or more annotators mention the gender in their description, and ``gender neutral'' otherwise.
We can then measure accuracy over these three classes.
Whereas a naive solution which restricts vocabulary to include no gender words would have low error as defined in Table~\ref{tab:accuracy-ratio}, it would not capture the way humans use gender words when describing images.
Indeed, the MSCOCO training set includes over 200,000 instances of words which describe people. %
Over half of all words used to describe people are gendered.
By considering accuracy across three classes, we can better measure how well models capture the way humans describe gender.

Figure~\ref{fig:confident_accuracy} plots the accuracy of each model with respect to the confidence threshold $T$.  At low threshold values, Equalizer performs worse as it tends to more frequently output gender neutral terms, and the UpWeight model, which almost always predicts gendered words, performs best.  
However, as the threshold value increases, Equalizer performs better than other models, including at a threshold value of $3$ which corresponds to classifying images based off the majority vote.
This indicates that Equalizer naturally captures when humans describe images with gendered or gender neutral words.

\myparagraph{Object Gender Co-Occurrence.}
We analyze how gender prediction influences prediction of other words on the MSCOCO-Bias test set.  Specifically, we consider the $80$ MSCOCO categories, excluding the category ``person''.
We adopt the bias amplification metric proposed in \cite{zhao2017men}, and
compute the following ratios: $\frac{count(man \& object)}{count(person \& object)}$ and $\frac{count(woman \& object)}{count(person \& object)}$, where \emph{man} refers to all male words, \emph{woman} refers to all female words, and \emph{person} refers to all male, female, or gender neutral words.
Ideally, these ratios should be similar for generated captions and ground truth captions.
However, e.g. for \emph{man} and \emph{motorcycle}, the ground truth ratio is 0.40 and for the Baseline-FT and Equalizer, the ratio is 0.81 and 0.65, respectively.
Though Equalizer over-predicts this pair, the ratio is closer to the ground truth than when comparing Baseline-FT to the ground truth.
Likewise, for \emph{woman} and \emph{umbrella}, the ground truth ratio is 0.40, Baseline-FT ratio is 0.64, and Equalizer ratio is 0.56.
As a more holistic metric, we average the \emph{difference} of ratios between ground truth and generated captions across objects (lower is better).
For male words, Equalizer is substantially better than the Baseline-FT (0.147 vs. 0.193) and similar for female words (0.096 vs. 0.99).

\begin{figure}[t]
\centering
\includegraphics[width=0.9\linewidth]{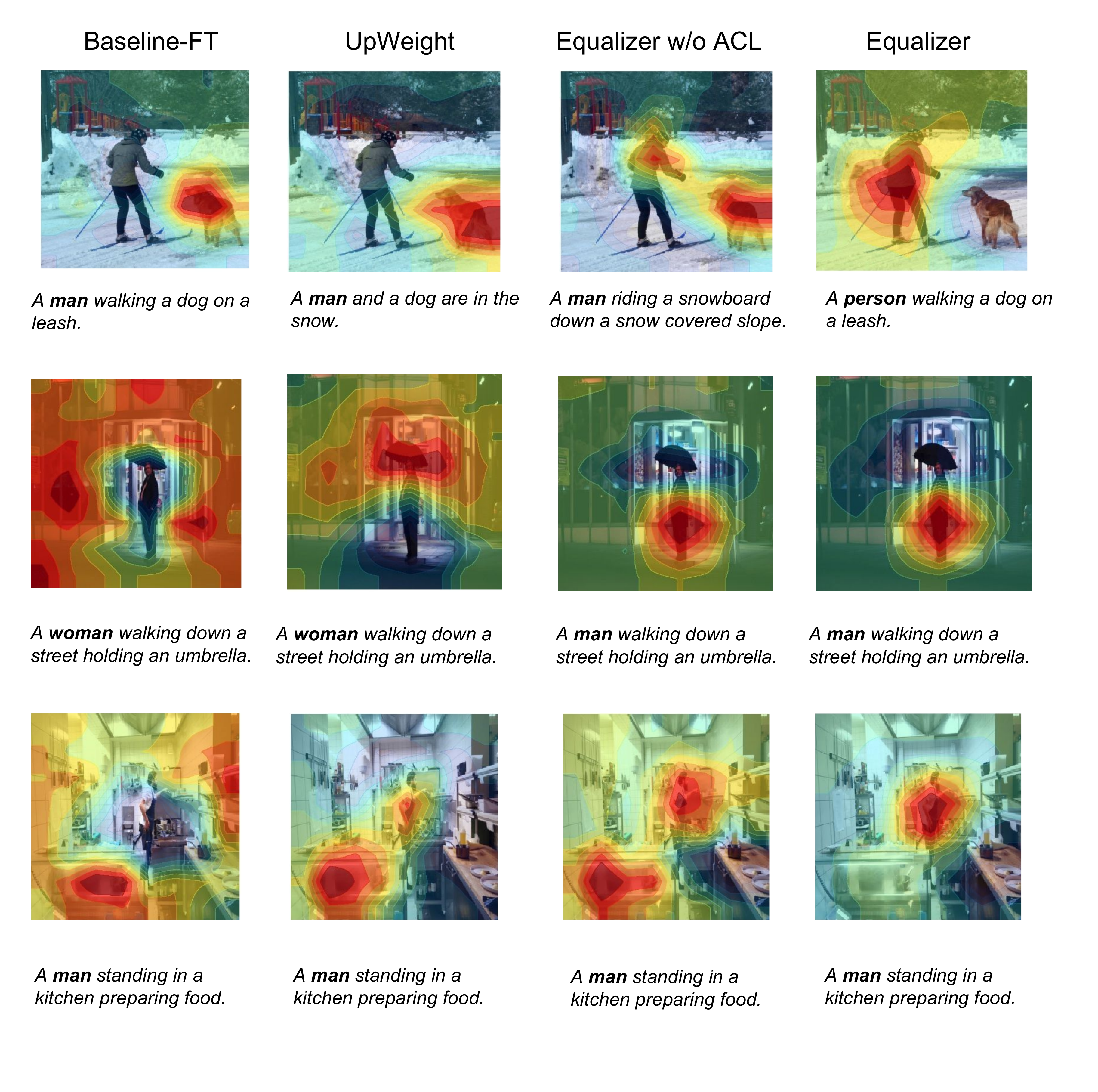}
\caption{\label{fig:pointing} Qualitative comparison of multiple baselines and our model. In the top example, being conservative (``person'') is better than being wrong (``man'') as the gender is not obvious. In the bottom example the baselines are looking at the wrong visual evidence.}
\end{figure}

\myparagraph{Caption Quality.} Qualitatively, the sentences from all of our models are linguistically fluent (indeed, comparing sentences in Figure \ref{fig:pointing} we note that usually only the word referring to the person changes).  However, we do notice a small drop in performance on standard description metrics (25.2 to 24.3 on METEOR~\cite{lavie2014meteor} when comparing Baseline-FT to our full \modelname{}) on MSCOCO-Bias.
One possibility is that our model is overly cautious and is penalized for producing gender neutral terms for sentences that humans describe with gendered terms.

\begin{table*}[t]
\centering
\begin{tabular}{ c @{\ \ \ \ \ } c }

\begin{tabular}{ l | c | c | c }
Accuracy  & Woman & Man  & All \\ 
  Random  & 22.6 & 19.5 & 21.0\\\midrule
  Baseline-FT & 39.8  & 34.3 & 37.0 \\ 
  Balanced & 37.6 & 34.1 & 35.8 \\
  UpWeight & 43.3 & 36.4 & 39.9 \\  
  \midrule
  \modelname{} w/o ACL & 48.1 & 39.6 & 43.8 \\
  \modelname{} w/o Conf  & 43.9 & 36.8 & 40.4 \\
  \modelname{} (Ours) & \textbf{49.9} & \textbf{45.2} & \textbf{47.5} \\
\end{tabular}
&

\begin{tabular}{ l | c | c | c }
Accuracy  & Woman & Man  & All \\ \midrule
  Random  & 25.1 & 17.5 & 21.3\\
  \midrule
  Baseline-FT & 45.3 & 40.4 & 42.8\\ 
  Balanced & 48.5 & 42.2 & 45.3 \\
  UpWeight & 54.1 & 45.5 & 49.8 \\  
  \midrule
  \modelname{} w/o ACL & 54.7 & 47.5 & 51.1 \\
  \modelname{} w/o Conf  & 48.9 & 46.7 & 47.8 \\
  \modelname{} (Ours) & \textbf{56.3} & \textbf{51.1} & \textbf{53.7} \\
\end{tabular} 

\\
\\
(a) Visual explanation is a \emph{Grad-CAM} map. & (b) Visual explanation is a \emph{saliency} map. \\
\\
\end{tabular}
\caption{\label{tab:pointing}\emph{Pointing game} evaluation that measures whether the visual explanations for ``man'' / ``woman'' words fall in the person segmentation ground-truth. Evaluation is done for ground-truth captions on the MSCOCO-Balanced.}
\end{table*}

\myparagraph{Right for the Right Reasons.}
We hypothesize that many misclassification errors occur due to the model looking at the wrong visual evidence, e.g. conditioning gender prediction on context rather than on the person's appearance. We %
quantitatively confirm this hypothesis and show that our proposed model improves this behavior by looking at the appropriate evidence, i.e. is being ``right for the right reasons''. To evaluate this we rely on two visual explanation techniques: Grad-CAM \cite{selvaraju2016grad}  and saliency maps generated by occluding image regions in a sliding window fashion. %

Unlike \cite{selvaraju2016grad} who apply Grad-CAM to an entire caption, we visualize the evidence for generating specific words, i.e. ``man'' and ``woman''. %
Specifically, we apply Grad-CAM to the last convolutional layer of our image processing network, InceptionV3 \cite{szegedy2016rethinking}, we obtain {8x8} weight matrices. 
To obtain saliency maps, we resize an input image to $299 \times 299$ and uniformly divide it into $32 \times 32$ pixel regions, obtaining a $10 \times 10$ grid (the bottom/rightmost cells being smaller). Next, for every cell in the grid, we zero out the respective pixels and feed the obtained ``partially blocked out'' image through the captioning network (similar to as was done in the occlusion sensitivity experiments in~\cite{zeiler2014visualizing}). Then, for the ground-truth caption, we compute the ``information loss'', i.e. the decrease in predicting the words ``man'' and ``woman'' as $-\text{log}(p(w_t = g_m))$ and $-\text{log}(p(w_t = g_w))$, respectively. This is similar to the top-down saliency approach of \cite{ramanishka2017top}, who %
zero-out all the intermediate feature descriptors but one. 

To evaluate whether the visual explanation for the predicted word is focused on a person, we rely on person masks, obtained from MSCOCO ground-truth person segmentations. We use the \emph{pointing game} evaluation \cite{zhang2016top}. We upscale visual explanations to the original image size. %
We define a ``hit'' to be %
when the point with the highest weight is contained in the person mask. The accuracy is computed as 
 $\frac{\#hits}{\#hits + \#misses}$.

Results on the MSCOCO-Balanced set are presented in Table \ref{tab:pointing} (a) and (b), for the Grad-CAM and saliency maps, respectively. %
For a fair comparison we provide all models with ground-truth captions. %
For completeness we also report the random baseline, where the point with the highest weight is selected randomly. We see that \modelname{} obtains the best accuracy, %
significantly improving over the {Baseline-FT} and all model variants. %
A similar evaluation on the actual generated captions shows the same trends. 

\myparagraph{Looking at objects.}
Using our pointing technique, we can also analyze which MSCOCO objects models are ``looking'' at when they \emph{do not} point at the person while predicting ``man''/``woman''. 
Specifically, we count a “hit” if the highest activation is on an object in question.
We compute the following ratio for each gender: number of images where an object is “pointed at” to the true number of images with that object. 
We find that there are differences
across genders, e.g. 
``umbrella'', ``bench'', ``suitcase''
are more often pointed at when discussing women, while e.g. 
``truck'', ``couch'', ``pizza''
– when discussing men. 
Our model reduces the overall ``delta'' between genders for ground truth sentences from an average 0.12 to 0.08, compared to the Baseline-FT. E.g. for ``dining table'' Equalizer decreases the delta from 0.07 to 0.03.

\myparagraph{Qualitative Results.}
Figure \ref{fig:pointing} compares Grad-CAM visualizations for predicted gender words from our model to the Baseline-FT, UpWeight, and Equalizer w/o ACL.
We consistently see that our model looks at the person when describing gendered words.
In Figure \ref{fig:pointing} (top), all other models look at the dog rather than the person and predict the gender ``man'' (ground truth label is ``woman'').
In this particular example, the gender is somewhat ambiguous, and our model conservatively predicts ``person'' rather than misclassify the gender.
In Figure \ref{fig:pointing} (middle), the Baseline-FT and UpWeight example both incorrectly predict the word ``woman'' and do not look at the person (women occur more frequently with umbrellas).
In contrast, both the \modelname{} w/o ACL and the \modelname{} look at the person and predict the correct gender.
Finally, in Figure \ref{fig:pointing} (bottom), all models predict the correct gender (man), but our model is the only model which looks at the person and is thus ``right for the right reasons.''

\myparagraph{Discussion.}
We present the \modelname{} model which includes an \lossnameapp{} to encourage predictions to be confused when predicting gender if evidence is obscured and the \lossnamecon{} which encourages predictions to be confident when gender evidence is present.
Our \lossnameapp{}, requires human rationales about what is visual evidence is appropriate to consider when predicting gender.
We stress the importance of human judgment when designing models which include protected classes.
For example, our model can use information about clothing type (e.g., dresses) to predict a gender which may not be appropriate for all applications. %
Though we concentrate on gender in this work, we believe the generality of our framework could be applied when describing other protected attributes, e.g., race/ethnicity and believe our results suggest \modelname{} can be a valuable tool for overcoming bias in captioning models.

\paragraph{Acknowledgements.}  This work was partially supported by
US DoD, the DARPA XAI program, and the Berkeley Artificial Intelligence Research (BAIR) Lab.

\clearpage
\section*{Supplemental}

\appendix

\section{Content}

This supplementary material provides additional quantitative and qualitative results to our main paper. The document is structured as follows.

Section \ref{sec:sup_analysis} provides detailed breakdown for per-word performance of the compared approaches, discusses models' behavior on the masked images, and provides results for training with a set of gendered words.

Section \ref{sec:sup_examples} shows more qualitative examples for the baselines and our model.

\section{\label{sec:sup_analysis}Additional analysis}%

\paragraph{Performance breakdown for biased words.}
We additionally analyze objects which co-occur with one gender more than the other.
For a careful analysis, we choose five words that are biased to co-occur with women (umbrella, kitchen, cell phone, table, and food) and five words which frequently co-occur with men (skateboard, baseball, tie, motorcycle, and snowboard). To choose biased words, we compute bias as is done in~\cite{zhao2017men} Section 3 for the most commonly occurring nouns ($>$ 250 times) in the MSCOCO-Bias training set.
We compute the error rate and the \textit{difference} between the ground truth ratio of women to men and the ratio produced by each captioning model, 
for images containing the above objects (Table~\ref{tab:biased_all}). We observe similar trends to our observations in the main paper.  
\modelname{} and \modelname{} w/o ACL have the lowest errors, with \modelname{} w/o ACL performing slightly better, suggesting the confidence term is important for low error rate.
Considering distance to the ground truth gender ratio, the \modelname{} model consistently outperforms other models.
One particularly interesting case study is the word ``kitchen'' in which the ground truth woman to man gender ratio is 0.946 (recall that the dataset contains a roughly 1:3 woman to man gender ratio, so a gender ratio close to 1.0 for a specific object suggests that a higher proportion of ``woman'' images include a ``kitchen'' than ``man'' images).
The \modelname{} model predicts a gender ratio of 1.0 (delta 0.054) whereas the next best model (\modelname{} w/o ACL) predicts a gender ratio of 0.806 (delta 0.14).  
The Baseline-FT model predicts a ratio of 0.586 (delta 0.361).

\begin{table}[t]
\centering
\resizebox{\textwidth}{!}{
\begin{tabular}{ l | p{1cm} | p{1cm} | p{1cm} | p{1cm} | p{1cm} | p{1cm} | p{1cm} | p{1cm} | p{1cm} | p{1cm} }
 & \rotatebox{90}{Umbrella} & \rotatebox{90}{Kitchen} & \rotatebox{90}{Cell Phone} & \rotatebox{90}{Table} & \rotatebox{90}{Food} & \rotatebox{90}{Skateboard} & \rotatebox{90}{Baseball} & \rotatebox{90}{Tie} & \rotatebox{90}{Motorcycle} & \rotatebox{90}{Snowboard} \\ %
 \midrule
 \multicolumn{11}{c}{Error} \\
 \midrule

 Baseline-FT &  0.303 &	0.277 &	0.172 &	0.200 &	0.154 &	0.028 &	\textbf{0.072} &	0.017	& 0.083	& 0.073	\\ %
  \modelname{} w/o ACL & 0.210 &	\textbf{0.157} &	0.145 &	\textbf{0.100} &	\textbf{0.085} &	\textbf{0.020} &	0.085 &	0.021 &	\textbf{0.054} &	\textbf{0.031}\\ %
 \modelname{} w/o Conf & 0.250  &	0.269 &	0.158 &	0.189 & 0.166	 &	0.028 &	0.038 &	 0.017 &	0.107 &	0.100	\\ %

 \modelname{} &  \textbf{0.176} &	0.181 &	\textbf{0.119} &	0.119 &	0.128 &	0.031 &	0.113 &	\textbf{0.011}	& 0.064	& 0.081	\\ %
\midrule
\multicolumn{11}{c}{$\Delta$ Ratio (Women:Men)} \\
\midrule
Baseline-FT & \textbf{1.074} & 0.358 & 0.278 & 0.351 & 0.213 & 0.011 & 0.004 & 0.013 & 0.103 & 0.094 \\ %
\modelname{} w/o ACL & 1.274 & 0.137 & 0.028 & 0.151& 0.092& 0.009 & \textbf{0.018} & \textbf{0.002} & 0.086 & 0.051 \\ %
\modelname{} w/o Conf &  1.931 & 0.291 & 0.212 & 0.275 & 0.127  & 0.008 & 0.023 & 0.015 & 0.084 & 0.081 \\ %
\modelname{} & 2.335 &  \textbf{0.057} & \textbf{0.009} & \textbf{0.131} & \textbf{0.031} & \textbf{0.001} & 0.046 & 0.008 & \textbf{0.077} & \textbf{0.033} \\ %

\end{tabular}}
\caption{\label{tab:biased_all} Breakdown of error rate and difference to ground-truth woman:man ratio over images with specific biased words. We see that the full \modelname{} generally outperforms the Baseline-FT. On error, \modelname{} w/o ACL performs best, followed by \modelname{}.  \modelname{} performs best when considering predicted gender ratio.}
\end{table}

\paragraph{Masked Images.}  We also consider the gender ratio when predicting sentences for masked images (images in the test set are masked in the same way as was done to train the \lossnameapp{} term).
Ideally, the ratio of predicted gender words should be close to $1.0$ on the masked images as gender information is obscured.
The man to woman gender ratios for the \modelname{} w/o ACL, \modelname{} w/o Conf, and \modelname{} are 3.45, 2.87, and 1.98 respectively (all other models have larger ratios than \modelname{} w/o ACL).
This suggests that again both our ACL loss and Conf loss are important for %
predicting a fair gender ratio when gender information is not present in the image.
Again, we achieve the best performance with our full \modelname{} model.

\paragraph{Training with a Set of Gendered Words.}
To achieve the results in the main paper, we only apply the Appearance Confusion Loss and Confident loss to sets \{man\} and \{woman\}. As shown in Table \ref{tab:accuracy-ratio-sets} %
, applying these losses to larger sets of gendered words, including man and woman, does not significantly affect performance on MSCOCO. The sets of gendered words used in this experiment are \{girl, sister, mom, wife, woman, bride, female, lady, women\} and \{boy, brother, dad, husband, man, groom, male, guy, men\} for women and men, respectively.

\begin{table*}[t]
\centering
\small
\begin{tabular}{ l@{\ \ }| r r | r r }
& \multicolumn{2} {c@{\ \ }@{\ \ }|} {MSCOCO-Bias}  & \multicolumn{2} {c} {MSCOCO-Balanced} \\
Model & Error & Ratio $\Delta$  & Error &  Ratio $\Delta$  \\
\midrule
  \midrule
Baseline-FT  & 12.83   & 0.15  & 19.30 &  0.51  \\
\modelname{} &  {7.02} & {-.03}   & {8.10} & \textbf{0.13}   \\
\modelname{} (Multiple Gender Words) &  \textbf{6.81} & \textbf{.00}   & \textbf{8.9} & {0.19}   \\
\end{tabular}
\caption{\label{tab:accuracy-ratio-sets} Evaluation of predicted gender words based on error rate and ratio of generated sentences which include the ``woman'' words to sentences which include the ``man'' words. Training \modelname{} with sets of multiple gendered words improves error and ratio on the MSCOCO-Bias set, but not significantly so.}
\end{table*}

\section{\label{sec:sup_examples}Qualitative Examples}

In the following we rely on Grad-CAM maps for visualization.

\begin{figure}[t]
\centering
\includegraphics[width=\linewidth]{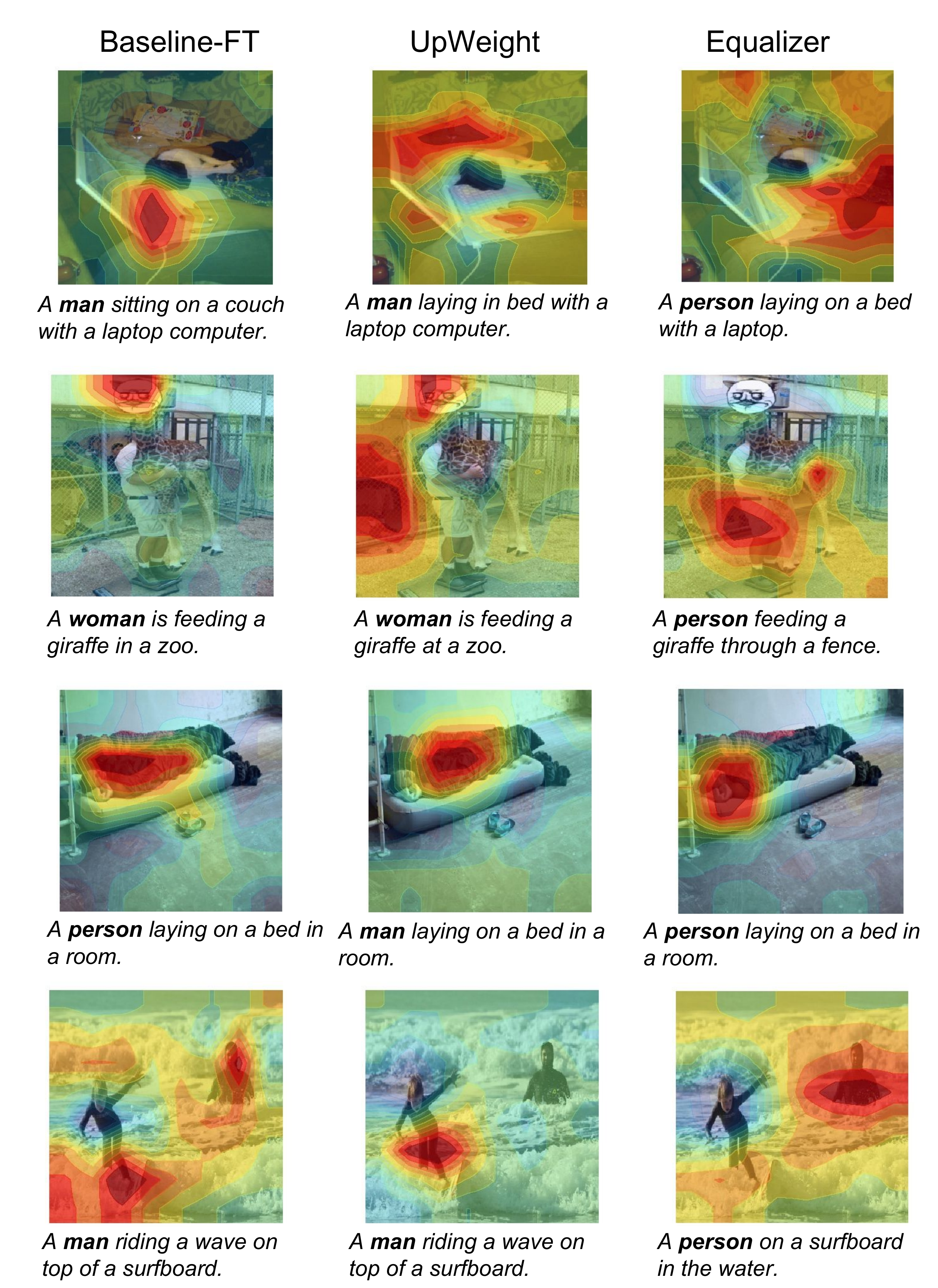}
\caption{\label{fig:sup_other} Qualitative comparison of baselines and our model when our model predicts ``person'' rather than ``woman'' or ``man''.}
\end{figure}

In Figure \ref{fig:sup_other} we provide multiple examples of images where our model \modelname{} predicts ``person'' rather than ``woman'' or ``man''. In many cases this occurs when the gender evidence is challenging (e.g. first example where only the person's hands and arms are visible and second example where the person's face is occluded by the giraffe) or the person's pose is unusual (third example). However, we also observe cases like the one at the bottom, where \modelname{} predicts ``person'' despite looking at the clear/correct gender evidence. We attribute this to the Confident Loss term, which allows for neutral words generation when the model is uncertain about gender.

\begin{figure}[t]
\centering
\includegraphics[width=\linewidth]{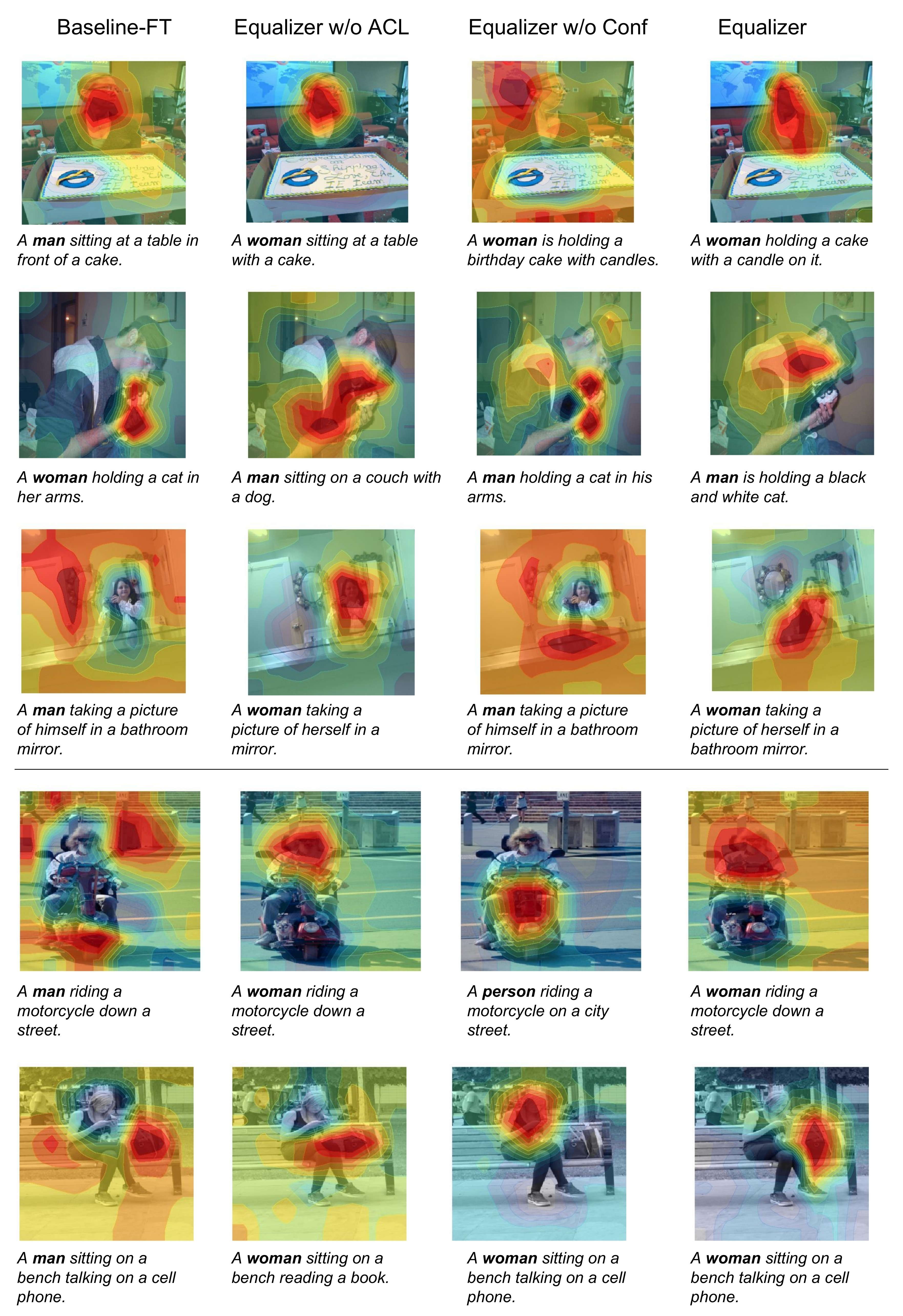}
\caption{\label{fig:sup_more} Qualitative comparison of baselines and our model. At the top we show success cases where our model predicts the right gender for the right reasons. At the bottom we show failure cases with incorrectly predicted gender and the wrong gender evidence.}
\end{figure}

Figure \ref{fig:sup_more} presents more qualitative examples for the baselines and our model. At the top we show success cases where our model predicts the right gender for the right reasons. At the bottom we show failure cases with incorrectly predicted gender and the wrong gender evidence.

\clearpage

\bibliographystyle{splncs}
\bibliography{egbib}
\end{document}